\documentclass{article}
\usepackage{ijcai20}

\usepackage{times}
\usepackage{soul}
\usepackage{url}
\usepackage[hidelinks]{hyperref}
\usepackage[utf8]{inputenc}
\usepackage[small]{caption}
\usepackage{graphicx}
\usepackage{amsmath}
\usepackage{amsthm}
\usepackage{amssymb}
\usepackage{amsfonts}
\usepackage{booktabs}
\usepackage{algorithm}
\usepackage{algorithmic}
\usepackage{multirow}
\usepackage{todonotes}
\usepackage{mathtools}
\usepackage{etoolbox}
\mathtoolsset{showonlyrefs}

\newcommand{\s}{\ensuremath{\textrm{Sim}}}

\newcommand{\dd}[1]{\ensuremath{d^{(#1)}}}
\newcommand{\ww}[1]{\ensuremath{w^{(#1)}}}

\newcommand{\vv}[1]{\ensuremath{v^{(#1)}}}

\DeclareMathOperator*{\argmin}{arg\,min}
\newtheorem{definition}{Definition}

\title{Structural-Aware Sentence Similarity with Recursive Optimal Transport}

\author{
    Anonymous Authors
    \affiliations
    Anonymous Affiliations
}

\author{
Zihao Wang$^{1,2,3}$
\and
Yong Zhang$^{1,2,3}$\and
Hao Wu$^4$
\affiliations
$^1$Research Institute of Information Technology\\
$^2$Beijing National Research Center for Information Science and Technology (BNRist)\\
$^3$Department of Computer Science and Technology, Tsinghua University, Beijing 100084, China\\
$^4$Department of Mathematical Sciences, Tsinghua University
\emails
wzh17@mails.tsinghua.edu.cn
\{zhangyong05,hwu\}@tsinghua.edu.cn,
}
\begin{document}

\maketitle

\begin{abstract}
 Measuring sentence similarity is a classic topic in natural language processing. 
 Light-weighted similarities are still of particular practical significance even when deep learning models have succeeded in many other tasks.
Some light-weighted similarities with more theoretical insights have been demonstrated to be even stronger than supervised deep learning approaches.
However, the successful light-weighted models such as Word Mover's Distance~\cite{DBLP:conf/icml/KusnerSKW15} or Smooth Inverse Frequency~\cite{DBLP:conf/iclr/AroraLM17} failed to detect the difference from the structure of sentences, i.e. order of words.
To address this issue, we present Recursive Optimal Transport (ROT) framework to incorporate the structural information with the classic OT.
Moreover, we further develop Recursive Optimal Similarity (ROTS) for sentences with the valuable semantic insights from the connections between cosine similarity of weighted average of word vectors and optimal transport.
ROTS is structural-aware and with low time complexity compared to optimal transport.
Our experiments over 20 sentence textural similarity (STS) datasets show the clear advantage of ROTS over all weakly supervised approaches.
Detailed ablation study demonstrate the effectiveness of ROT and the semantic insights.
\end{abstract}

\section{Introduction}

Measuring sentence similarity is one of the central topics in natural language processing.
Early attemptations are based on graphical model~\cite{DBLP:journals/jmlr/BleiNJ03} that did not take the semantic similarity into account.
After the semantic information could be embedded into Euclidean space~\cite{DBLP:conf/nips/MikolovSCCD13,DBLP:conf/emnlp/PenningtonSM14}, neural networks~\cite{DBLP:conf/icml/LeM14,DBLP:conf/nips/KirosZSZUTF15,DBLP:conf/emnlp/ConneauKSBB17} are proposed to model sentence similarity by supervised learning. 
However, those supervised deep learning approach are showed~\cite{DBLP:conf/iclr/WietingBGL15a} to be easily overfitted, especially evaluated in transfer learning settings.
Pre-trained Language Models (PLM) such as BERT~\cite{DBLP:conf/naacl/DevlinCLT19} succeed in many natural language tasks.
However, building and using the those PLMs are extremely data consuming and computationally expensive.
Hence, seeking for light-weighted sentence similarities with excellent performance are still of particular practical interests.

Here, the term ``light-weighted'' limits the amount of required pre-trained information as well as the computing power, and excludes any training and finetuning.
Such models could be used in low-performance devices and varying scenarios.
Recently, two kinds of light-weighted similarities are frequently discussed.
They are only based on pre-trained word vectors $v_{(\cdot)}^{\cdot}$ and weights $w_{(\cdot)}^{\cdot}$ from the statistics of word occurrence .
One kind is Optimal Transport (OT) based approach~\cite{DBLP:conf/icml/KusnerSKW15,DBLP:conf/nips/HuangGKSSW16,DBLP:conf/emnlp/WuYXXBCRW18} and the other is cosine similarity of weighted average of word vectors (WAWV)~\cite{DBLP:conf/iclr/AroraLM17,DBLP:conf/rep4nlp/EthayarajhH18}.

Word Mover's Distance (WMD)~\cite{DBLP:conf/icml/KusnerSKW15} employed the Wasserstein metric to measure the distance of sentences by treating them as normalized bag of words (nBOW) distributions in word embedding spaces.
Given two sentences $\dd{1} $ and $\dd{2} $ with word vectors $\{\vv{1}_i\}_{i=1}^m, \{\vv{2}_j\}_{j=1}^n$ with nBOW weights $ \{\ww{1}_i\}_{i=1}^m,\{\ww{2}_j\}_{j=1}^n$,  the WMD of $\dd{1}$ and $\dd{2}$ is:
\begin{align}
\mathrm{WMD}(d_1,d_2) = \min_{\Gamma\in\mathbb{R}^{m\times n}}\sum_{ij} \Gamma_{ij}D_{ij},\label{eq:kp}\\
\text{ s.t. } \sum_j \Gamma_{ij} = \ww{1}_i, \sum_i \Gamma_{ij} = \ww{2}_j.
\end{align}
where $D_{ij} = D(\vv{1}_i, \vv{2}_j)$ is the transport cost of two word vectors.
Equation~\eqref{eq:kp} minimizes the transport cost amongst the word embeddings.
With optimal transport, WMD explains how the sentence distance could be decomposed into word distances by the transport plan $\Gamma$~\footnote{previous OT based approach produces the sentence distances rather than similarities. But we note that the sentence distance and similarity could be easily converted to each other. For example, similarity~\eqref{eq:doc-cos-sim} is a bounded symmetry function $\s(\cdot, \cdot)\in [0, 1]$, we could get the distance by $1-\s(\cdot, \cdot)$.}.

Arora et al,\shortcite{DBLP:conf/iclr/AroraLM17} proposed to model the sentence $d$ by WAWV $ d = \sum_i w_i v_i $.
They argue that the cosine similarity of two sentence vectors in Equation~\eqref{eq:doc-cos-sim}  could be a simple but surprisingly effective similarity measurement.
\begin{align}\label{eq:doc-cos-sim}
\s(\dd{1}, \dd{2}) & = \frac{\langle \dd{1}, \dd{2} \rangle}{\lVert \dd{1} \rVert \lVert \dd{2} \rVert}.
\end{align}
WAWV with SIF~\cite{DBLP:conf/iclr/AroraLM17} weight scheme and its unsupervised variation uSIF~\cite{DBLP:conf/rep4nlp/EthayarajhH18} explains a sentence by the log-linear random walk and the cosine similarity between sentence vectors clearly outperform deep neural network models.

\begin{figure}[t]
	\centering
	\includegraphics[width=\linewidth]{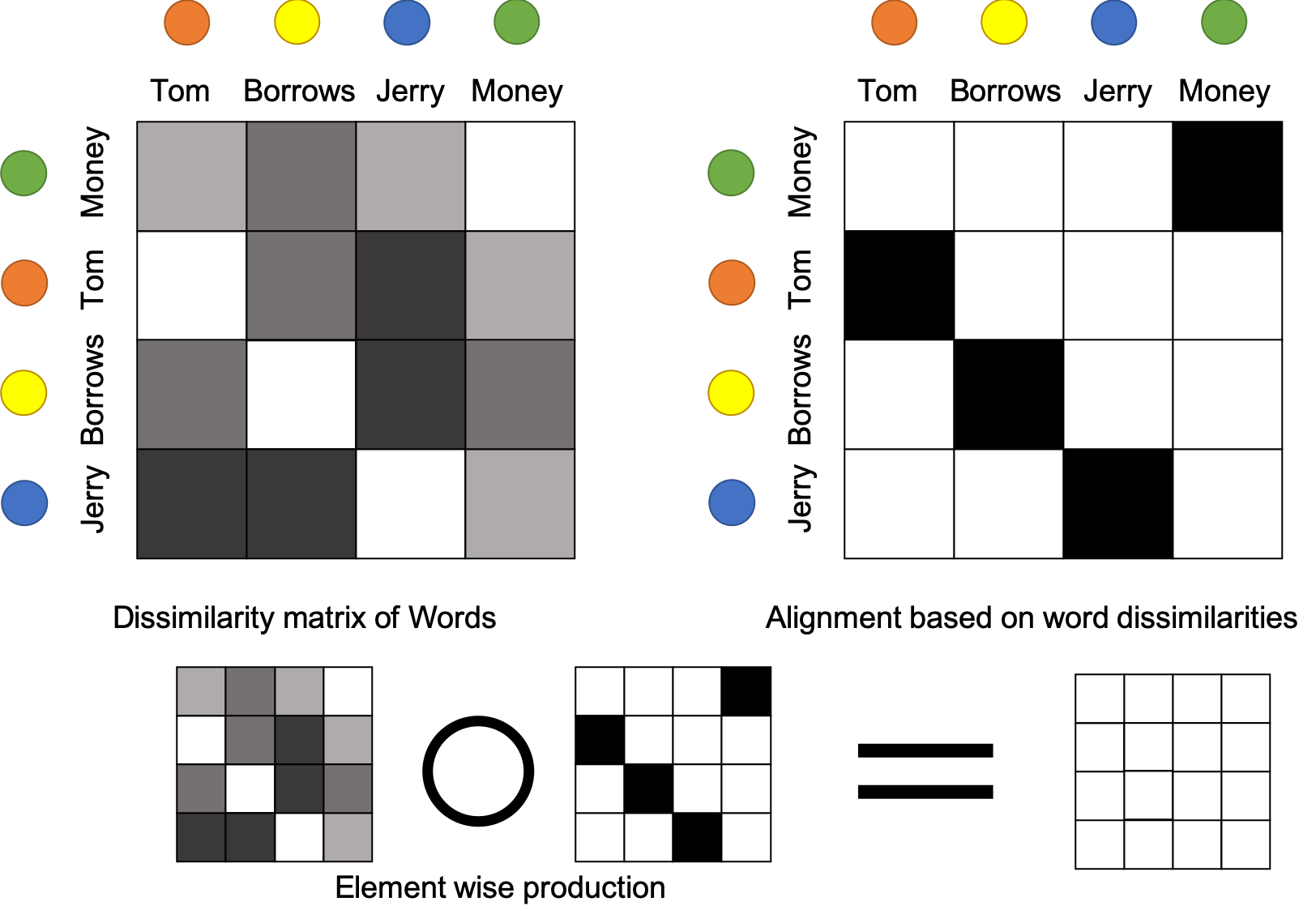}
	\caption{Optimal transport distance fails to discriminate the differences brought by the order of words. White blocks in the matrix indicate zero.}
	\label{fig:difficulty-of-ot}
\end{figure}

Though simple, neither cosine similarity of WAWV nor OT distance is able to capture the difference from structure of sentences.
Considering the following two different sentences:
\begin{itemize}
    \item[A] \textit{Tom borrowed Jerry money}
    \item[B] \textit{Jerry borrowed Tom money}
\end{itemize}
WAWV produces exactly the same sentence vectors due to the vector addition is communicative.
So the cosine similarity of WAWV is 1.
OT distance like WMD must be zero due to the optimal property of its solution demonstrated in Figure~\ref{fig:difficulty-of-ot}).
But obviously, sentence A and B are different because of the order of ``Tom'' and ``Jerry''.
Such structural information like word order is sometimes important in sentence similarity tasks.
More generally, let's consider a tree structure in the sentences where each word are related to only one the nodes in the tree, e.g. the dependency parsing trees.
We emphasize that the difference appears in the example could be eventually reflected in the tree structure.

To capture the structure information, we propose Recursive Optimal Transport (ROT), a novel framework to compare the distributional data under certain tree structural information.
Furthermore, we establish the semantic insights from the connection between OT and cosine similarity of WAWV.
With those insights, we adapt the ROT framework into a structural-aware sentence similarity named by Recursive Optimal Transport Similarity (ROTS).

The contributions of this paper are three folds:
\begin{itemize}
	\item We present Recursive Optimal Transport (ROT), a novel framework for comparing distributional data based on the consecutive solutions of the KL-regularized Optimal Transport (KOT) Problems under the structural information. ROT framework admits any internal or external tree-structures, where we could insert our prior hierarchical understanding of the data.
	\item We establish the semantic insights from the connection between OT and cosine similarity of WAWV. This connection brings us additional inspiration for the original ROT framework, which is essential for Recursive Optimal Transport Similarity (ROTS), the proposed structural-aware sentence similarity.
	\item We conduct extensive experiments on sentence similarity tasks with different pre-trained word vectors and tree structures. The results showed that ROTS reached clearly better performance than all other OT or WAWV based approaches. The complexity of ROTS is $O(n)$ rather than $O(n^2)$ in usual optimal transport, where $n$ is the length of sentences, which allows ROTS to scale up to large applications.
\end{itemize}

\section{Related Works}

\noindent\textbf{Word Embeddings}
Word embeddings embeds the words into the continuous Euclidean space~\cite{DBLP:conf/nips/MikolovSCCD13}. Compared to traditional one-hot embedding, the word vectors reflect the semantic relationship in metric space. For example, similar words are close in some norm (e.g., $ \lVert v_{\textrm{good}} - v_{\textrm{nice}} \rVert < \lVert v_{\textrm{good}} - v_{\textrm{Paris}} \rVert $) and some relationship could be captured by algebraic operation (e.g. $ \lVert (v_{\textrm{U.S.}} - v_{\textrm{New York}}) - (v_{\textrm{France}} - v_{\textrm{Paris}}) \rVert\approx 0  $). Word embeddings are trained from either local co-occurrence~\cite{DBLP:conf/nips/MikolovSCCD13} in a context window or low-rank approximation of global co-occurrence matrices~\cite{DBLP:conf/emnlp/PenningtonSM14}. Word vectors are light-weighted pre-trained information and could be looked up within $O(1)$ in hash tables. Our framework are based on distribution over word embedding spaces.

\noindent\textbf{Sentence Similarity}
Early approaches~\cite{DBLP:journals/jmlr/BleiNJ03} used inferred latent variables in the graphical model to represent the document. They are lack of the semantic information between words.
Based on word embeddings, some neural network composed the word embeddings into the sentence vectors by either direct mean~\cite{DBLP:conf/icml/LeM14} or neural networks~\cite{DBLP:conf/nips/KirosZSZUTF15} and under transfer learning setting~\cite{DBLP:conf/emnlp/ConneauKSBB17}. The sentence similarity is naturally defined by cosine similarity.

Recently, the weighted average of word vectors as sentence embedding~\cite{DBLP:conf/iclr/AroraLM17,DBLP:conf/rep4nlp/EthayarajhH18} achieved great success in semantic textual similarity tasks with surprising simplicity, better performance than previous deep models and clear theory explanation. They are inspired by~\cite{DBLP:conf/iclr/WietingBGL15a}.
By modeling sentence as a random walk process, the weight scheme is related to the statistics in the corpus and have no parameters to learn. 
SIF~\cite{DBLP:conf/iclr/AroraLM17} consists of two steps: (1) compute the weighted average of word vectors and (2) substract from each vector the projection on the first principle component among all vectors in the corpus. As an extension, uSIF~\cite{DBLP:conf/rep4nlp/EthayarajhH18} estimates the hyper-parameter in SIF and take a three-steps approach named by normalize-average-substraction to generate better sentence vector. 
Our work is related to the cosine similarity of WAWV but not relies on any specific form of weight scheme or word vectors.

\noindent\textbf{Optimal Transport for Sentence Similarity}
Optimal transport has also been applied as the sentence dissimilarity~\cite{DBLP:conf/icml/KusnerSKW15}.
And very soon, the application of supervised metric learning~\cite{DBLP:conf/nips/HuangGKSSW16} and Monte-Carlo embedding with linear-kernel SVM classifier~\cite{DBLP:conf/emnlp/WuYXXBCRW18} are developed.
Our work is also related to optimal transport. But compared to previous classic OT based approach that applied Wasserstein distance directly, we propose Recursive Optimal Transport framework. 
ROT framework enables the model includes structural information and other semantic insights.

\section{Methods}

In this part, we introduce the Recursive Optimal Transport, a framework that extends classical optimal transport with structural information.
In our paper, we mainly focused on the sentence $d $ with weights $\{w_i\}$ and vectors $\{v_i\}$, but we emphasis that it is trivial to apply ROT to more general cases where we handle the empirical distribution $P_d = \sum_{i} w_i \delta_{v_i}$.
$\delta$ here is the Dirac function.
In more general cases, the Weighted Average of Word Vector is actually to the expectation of the empirical distribution.
We also assume the tree structure $\mathcal{T}$, where $\mathcal{T}_k$ is the set of nodes at $k$-th tree level.
Node $S_{j}^{(k)} \in T_k$ is the $j$-th node at the $k$-th level $T_k$.
Each node $S_{j}^{(k)}$ is also considered as a \textbf{substructure set} that contains all words in the sub-tree whose root is $S_{j}^{(k)}$.
For all substructures $S_{j}^{(k)}$ at the same level $\mathcal{T}_k$, we have $ S_{i}^{(k)} \cap S_{j}^{(k)} = \emptyset $ and $ \cup_i S_{i}^{(k)} = \{\text{all words in the sentence}\} $.
Finally, we present Recursive Optimal Transport Similarity (ROTS), where we combine ROT with the semantic insights derived by the connection of cosine similarity of WAWV and OT.

\subsection{Tree Structure and Weighted Average}\label{sec:tree}

Let $ d = \sum_i w_i v_i $ to be the sentence by weighted average of word vectors, and  $ \{S_j\} $ to be substructures at any level of $\mathcal{T}$, we suggest to use weighted average to represent substructure $S_j$.
By doing this, the sentence vector could be represented again as \textbf{Weighted Average of Substructure Vectors} at the same level of $\mathcal{T}$.
\begin{align}
d & = \sum_i w_i v_i = \sum_j \sum_{i \in S_j} w_i v_i \\
& = \sum_j \left(\sum_{i \in S_j} w_i\right) \left(\sum_{i \in S_j} \frac{w_i}{\sum_{k\in S_i} w_k} v_i\right)\\
& = \sum_j \tilde{w}_j \tilde{v}_j.
\end{align}
The substructure $S_j$ are related to the weight $ \tilde{w}_j = \sum_{i\in S_j} w_i $ and the vector $ \tilde{v} = \sum_{i \in S_j} w_j v_j/\tilde{w}$. With this notation, the form of weighted average is persevered.

The substructure itself could be \emph{recursively} represented by the weighted average of vectors of its sub-sub-structures. This fact makes it particularly suitable to the hierarchical tree structures in sentences.
In this paper, we consider the binary tree structures for intrinsic order and dependence parsing tree as an example of external knowledge. In the binary tree case, the words are divided into two halves by their positions in sentences. Each half forms a sub-structures of sentences and then is \emph{recursively} divided into two halves till the we get single words.

\begin{figure}[t]
	\centering
	\includegraphics[width=0.9\linewidth]{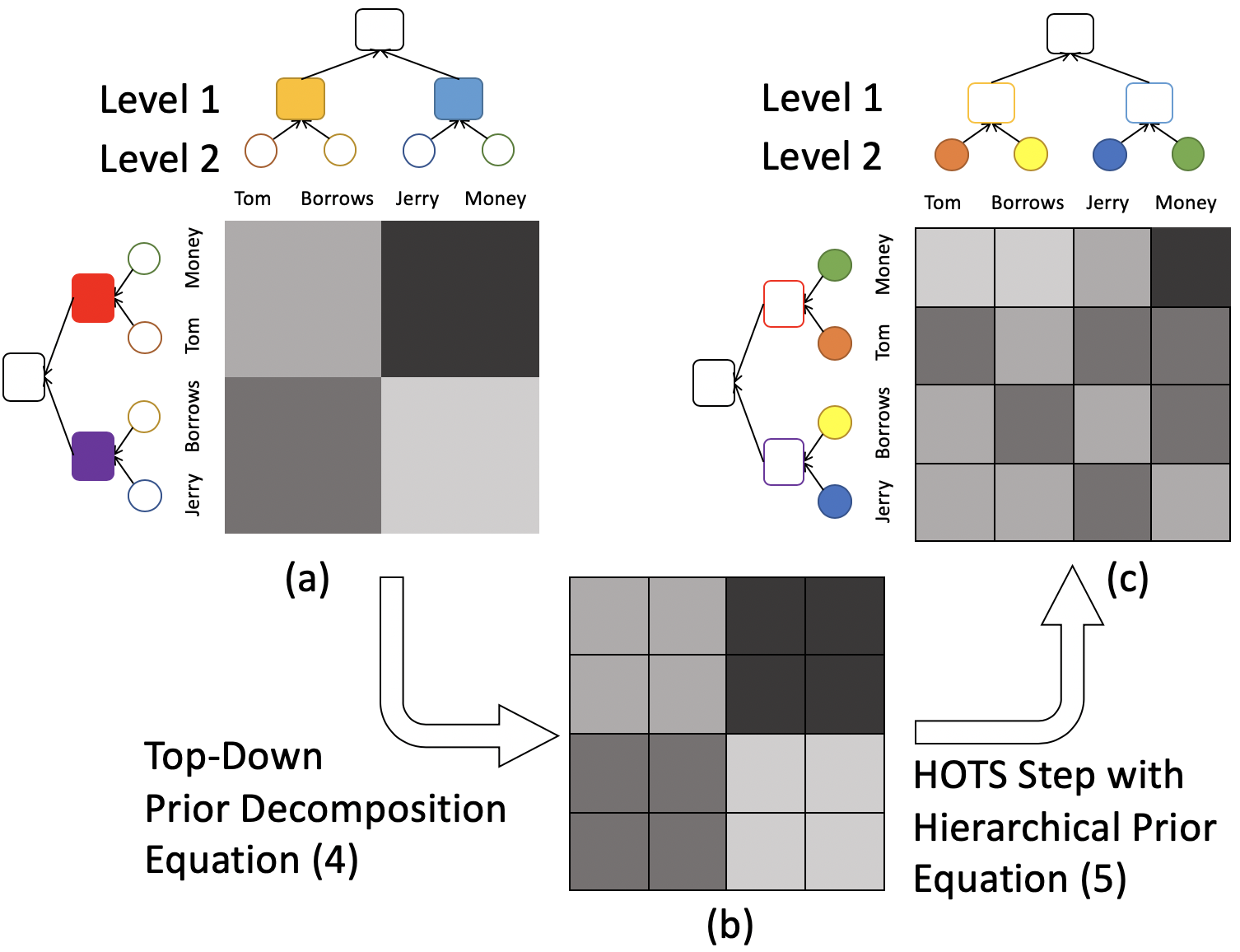}
	\caption{An Example of Recursive Optimal Transport. (a) Binary tree hierarchy of sentences is sensible to the order of words; (b) Level 1 alignment is decomposed as the prior for level 2; (c) Level 2 solution with hierarchical prior.}\label{fig:hots}
\end{figure}

As we discussed, the most important advantage of such tree structure is that the order of words could be detected. In the previous example, when we consider those two sentences A and B with binary tree structure, the first level of two sentences are A: \textit{ \{Tom borrowed\} \{Jerry money\} } and B: \textit{\{Jerry borrowed\} \{Tom money\}}. The lost difference in either sentence cosine similarity of WAWV or OT is re-discovered in Figure~\ref{fig:hots} (a).

\subsection{Recursive Optimal Transport}

The sentence is recursively breakdown under some tree structure $\mathcal{T}$.
At each level $\mathcal{T}_k$, we have the weights $\{w_{j,(k)}\}$ and vectors $\{v_{j,(k)}\}$ of the substructures.
So we could compute optimal transport at each level exactly the same as Word Mover's distance and obtain the transport plan $T$. Transport plan $T$ at different level explain how the different substructures align with others.

It is natural to conduct the computation process of optimal transport based on the tree structure. We suggest the computation is done by top-down order and the information is passed from higher-level to lower-level.
This kind of formulation have two advantages: 
(1) The optimal transport at each level will produces the alignment with different abstraction. This brings us structural aware alignment whose effect is demonstrated in the example 
(2) Higher-level~\footnote{In our definition, the level is higher when it is closed to the root} alignment could guide the alignment of lower level. When the computation goes deeper, the transport plan becomes more detailed but still preserves important information from higher-level. This makes the alignment solution at the lower-level more suitable reflects the structural information from higher-level.
With the second unique feature, we denote our framework as Recursive Optimal Transport.

We use the concept ``prior'' to pass the structural information from the higher-level alignment to lower-level solution.
More specifically, we propose KL-regularized Optimal Transport (KOT) to obtain lower-level solution based some certain prior information.

\begin{definition}[KL-regularized Optimal Transport]
	Given two documents $ \dd{1} = \sum_i \ww{1}_i \vv{1}_i $ and $ \dd{2} = \sum_i \ww{2}_i \vv{2}_i $, where $ \ww{\cdot} $ are weights and $ \vv{\cdot} $ are vectors for sub-structures at some level and prior $ T_p $. The transport plan is the solution $ \Gamma^* $ by the following minimization problem
	\begin{align}\label{eq:kl-reg-ot}
	\Gamma^* = \argmin_{\Gamma} & \sum_{ij} \Gamma_{ij} (1- \s(\vv{1}_i, \vv{2}_j)) + \gamma_p D_{KL}(\Gamma \|  T_p), \\ 
	\text{ s.t. } & \sum_i \Gamma_{ij} = \ww{2}_j, \sum_j \Gamma_{ij} = \ww{1}_i,
	\end{align}
	where we use $1-\s(\cdot, \cdot)$ to present the distance between two vectors.
	The solution $\Gamma^*$ could be used to calculate the distance of two sentences by $\textrm{Dist}(\dd{1}, \dd{2}) = \sum_{ij} (1- \s(\vv{1}_i, \vv{2}_j))\Gamma^*_{ij}$.
\end{definition}
Compare to Equation~\eqref{eq:kp}, KOT inserts an additional KL-regularization term in Equation~\eqref{eq:kl-reg-ot}. This term allows us to insert the external information.
We note that the KL regularization could be regarded as linear term plus the entropy.
\begin{align}
\sum_{ij} \Gamma_{ij} &(1- \s(\vv{1}_i, \vv{2}_j)-\gamma_p\log T_{p, ij})+ \gamma_p H(\Gamma).
\end{align}
Then it could be solved by Sinkhorn iteration~\cite{DBLP:conf/nips/Cuturi13}. In our implementation, we do not require the $ \gamma_p $ close to zero to obtain the approximated classical optimal transport results like all other OT based models~\cite{DBLP:conf/icml/KusnerSKW15,DBLP:conf/nips/HuangGKSSW16,DBLP:conf/emnlp/WuYXXBCRW18}. This allows very fast convergence of the Sinkhorn iteration as well as totally different solutions. 
Here, $ \gamma $ is the hyper-parameter controls the ``interpolation'' between the optimal transport and the \textit{transport under certain transport prior}. It is noticed that when $ \gamma_p \to 0 $, the solution $ \Gamma^* $ of~\eqref{eq:kl-reg-ot} converges to optimal transport plan and the $ \textrm{Dist}(\dd{1}, \dd{2}) \to \textrm{WMD}(\dd{1}, \dd{2}) $. When $ \gamma_p \to \infty $, the $ \Gamma^* \to T_p $  and $ \textrm{Dist}(\dd{1}, \dd{2}) $ converges to the distance by prior transport plans.
We also note that the regularization term could be more than one. Multiple KL-regularization terms could be breakdown into linear terms and entropy terms. There is no non-trivial difficulty to solve multiple KL-regularizations with Sinkhorn iteration.

After obtained the $ k $-th level KOT transport plan $ \Gamma^{(k)} $, we split $ \Gamma^{(k)} $ into $ T^{(k+1)} $ as the prior for next-level KOT.
Given $ \Gamma^{(k)}_{ij} $ the mass that transports from $ i $-th substructure $ S_i $ to $ S_j $, we consider the word index $ m_i\in S_i $ and $ n_j \in S_j $, then the prior transport plan $ T^{(k+1)} $ is defined by
\begin{align}\label{eq:hot-weight-split}
T^{(k+1)}_{m_i n_j} = \Gamma^{(k)}_{ij} \frac{\tilde{w}_{m_i,(k)}\tilde{w}_{n_j,(k)}}{\sum_{ \tilde m_i \in S_i, \tilde n_j \in S_j}\tilde{w}_{\tilde m_i,(k)}\tilde{w}_{\tilde n_j,(k)}}.
\end{align}
As shown in Figure~\ref{fig:hots} (b), Equation~\eqref{eq:hot-weight-split} generated $ T^{(k+1)} $ as a ``refinement'' of $ \Gamma^{(k)} $.
In this way, we recursively breakdown the previous solution of higher-level optimal transport into lower-level transport prior.
With the top-down prior splitting scheme in Equation~\eqref{eq:hot-weight-split}, solving  the following problem~\eqref{eq:rot-step} brings us structural-aware solution for the alignment at next level.
\begin{align}
\min_{\Gamma} \sum_{ij} \Gamma_{ij} (1- &\s(\vv{1}_{i,(k+1)}, \vv{2}_{j,(k+1)})) + \gamma_k D_{KL}(\Gamma \|  T^{(k+1)}),\\
\text{ s.t. } \sum_i \Gamma_{ij}& = \ww{2}_{j,(k+1)}, \sum_j \Gamma_{ij} = \ww{1}_{i,(k+1)}.\label{eq:rot-step} 
\end{align}

Take the binary tree structure as an example, we first solve the 2-to-2 KOT problem, its resulting 2-to-2 transport plan $\Gamma^{(1)}$ could be downward split as the prior $T^{(2)}$ for 4-to-4 KOT problem, and then 8-to-8, 16-to-16 KOT problems and etc. 
In our implementation, we find that use transport plans of 1-5 levels are sufficient. This means that we only need to handle Sinkhorn iteration up to 32-to-32 scale. This makes our algorithm scalable to very long sentences. Finally, we ensemble the similarities from all levels to obtain the ROT distance.
The process described above is called Recursive Optimal Transport shown in Algorithm~\ref{algo:rot} .
\begin{algorithm}
\caption{Recursive Optimal Transport}\label{algo:rot}
    \begin{algorithmic}
        \REQUIRE Sentence $\dd{1} = \sum \ww{1}_i \vv{1}_i$, $\dd{2}= \sum \ww{2}_i \vv{2}_i$, Tree structure $\mathcal{T}^{(1)}$ and $\mathcal{T}^{(2)}$. ROT depth $d$ and regularization $\gamma_k$
        \ENSURE Recursive Optimal Transport Distance $ROT$
        \STATE Compute the first-level transport plan $\Gamma^{(1)}$ by Equation~\eqref{eq:kp}
        \FOR {$k \gets 2,...,d$}
        \STATE Compute $k$-th level prior transport plan $T^{k}$ using $\Gamma^{k-1}$ by Equation~\eqref{eq:hot-weight-split}.
        \STATE Compute $k$-th level transport plan $\Gamma^{(k)}$ by Equation~\eqref{eq:rot-step}
        \ENDFOR
        \STATE $ROT\gets \frac{1}{d} \sum_{k=1}^d \sum_{ij} \Gamma_{ij}^{(k)} (1- \s(\vv{1}_{i,(k)}, \vv{2}_{j,(k)}))$
    \end{algorithmic}
\end{algorithm}

\subsection{New Semantic Insights by Cosine Similarity}\label{sec:insights}
Previous OT solution connects the word distance with sentence distance.
The distance is positive and not bounded.
However, when it comes to sentence similarity, especially bounded similarity case, we find that directly applying optimal transport misses the semantic insights by cosine similarity of WAWV.
We establish the connection between cosine similarity of WAWV and OT as follows: 
\begin{align}
\s(\dd{1}, \dd{2}) &= \frac{1}{\lVert \dd{1} \rVert \lVert \dd{2} \rVert} \langle \sum_i \ww{1}_i \vv{1}_i, \sum_j \ww{2}_j \vv{2}_j \rangle\\
& = C \sum_{ij} T_{ij} \s(\vv{1}_i, \vv{2}_j),\label{eq:cos-sim-meets-ot}
\end{align}

where 
\begin{align}
C &= \frac{\sum_k \ww{1}_k\lVert\vv{1}_k\rVert \sum_k \ww{2}_k\lVert\vv{2}_k\rVert}{\lVert \dd{1} \rVert \lVert \dd{2} \rVert},
T_{ij} = a_i b_j,\label{eq:notation1}\\
a_i &= \frac{\ww{1}_i\lVert\vv{1}_i\rVert}{\sum_k \ww{1}_k\lVert\vv{1}_k\rVert},
b_j = \frac{ \ww{2}_i\lVert\vv{2}_i\rVert}{\sum_k \ww{2}_k\lVert\vv{2}_k\rVert}\label{eq:notation2}
\end{align}

We emphasis that the formulation in equation~\eqref{eq:cos-sim-meets-ot} is similar to the Kantorovich formulation~\eqref{eq:kp} of the optimal transport theory except the constant $ C $ and separable $T$.
Despite of optimal $ \Gamma^* $ in Equation~\eqref{eq:kp}, $ T $ here is the suboptimal ``transport plan''. 
In this way, cosine similarity could be viewed as a transportation process based on the induced transport plan $ T $ and word similarity matrix $ \s(\vv{1}_i, \vv{2}_j) $. 



Interestingly, coefficient $C$ induced by cosine similarity brings us novel semantical insights. Let's consider
$C = \frac{\sum_k \ww{2}_k\lVert\vv{2}_k\rVert}{\lVert\sum_k \ww{2}_k\vv{2}_k\rVert} \frac{\sum_k \ww{1}_k\lVert\vv{1}_k\rVert}{\lVert\sum_k\ww{1}_k\vv{1}_k\rVert} = \sqrt{K_1 K_2} $. We note that for specific $K_{\cdot}$ we have
\begin{align}
&K_{\cdot} = \frac{(\sum_k \ww{\cdot}_k\lVert\vv{\cdot}_k\rVert)^2}{\lVert\sum_k \ww{\cdot}_k\vv{\cdot}_k\rVert^2} \\
= & \frac{\sum_{k} (\ww{\cdot}_k)^2 \lVert \vv{\cdot}_k \rVert ^ 2 + \sum_{k\neq m} \ww{\cdot}_k \ww{\cdot}_m \lVert \vv{\cdot}_k\rVert \lVert \vv{\cdot}_m\rVert}{\sum_{k} (\ww{\cdot}_k)^2 \lVert \vv{\cdot}_k \rVert ^ 2 + \sum_{k\neq m} \ww{\cdot}_k \ww{\cdot}_m \langle \vv{\cdot}_k, \vv{\cdot}_m \rangle}\\
= & 1 + \sum_{k\neq m} \frac{\ww{\cdot}_k \ww{\cdot}_m \lVert \vv{\cdot}_k\rVert \lVert \vv{\cdot}_m\rVert}{\lVert \dd{\cdot} \rVert^2} (1 - \s(\vv{\cdot}_k, \vv{\cdot}_m))\label{eq:k},
\end{align}
where the second summation term in Equation~\eqref{eq:k} represents the degree of intra-sentence word distance (1 - cosine similarity). So the factor $ C=\sqrt{K_1 K_2} $ encourages that the sentence pair containing sentences range from many semantic details to be more similar. 
Notably, some minor semantic outliers will disturb the document similarity when minimizing $ \sum_{ij} \Gamma_{ij} (1 - \s(\vv{\cdot}_k, \vv{\cdot}_m)) $ optimal transport and reduce the similarity unexpectedly. Comprehensive effect of $ C $ and $ \sum_{ij} \Gamma_{ij} (1 - \s(\vv{\cdot}_k, \vv{\cdot}_m)) $ is the key missing part of pure OT based sentence distance.
Also, $a_i$ and $b_j$ are modified weight schemes with word vectors. Compared to original weight, $a_i$ and $b_j$ are emphasizing the words whose embedding has larger norm. This makes it easier to recognize the important words.


\subsection{Recursive Optimal Transport Similarity}

With those semantic insights by cosine similarity, we Propose Recursive Optimal Transport Similarity (ROTS).
Compared to ROT shown in algorithm~\ref{algo:rot}, we have three difference:
\begin{description}
    \item[From distance to similarity]Instead calculating the distance, the transport plan $\Gamma$ from Equation~\eqref{eq:kp} and \eqref{eq:rot-step} is used to derive the similarity of two sentences.
    \item[Modified Weight Scheme] Instead use the weights $\ww{\cdot}$ directly, we use $a$ and $b$ in Equation~\eqref{eq:notation2} to combine the word vector strength with the weights.
    \item[Comprehensive Coefficient] The similarity by $\Gamma$ is further multiplied by $C$ to produce the final similarity $\s(\dd{1}, \dd{2}) = C \sum_{ij} \s(\vv{1}_i, \vv{2}_j)\Gamma^*_{ij}$.
\end{description}
To the end, we have ROTS as
\begin{equation}
     ROTS = \frac{1}{d} \sum_{k=1}^d C^{(k)} \sum_{ij} \Gamma_{ij}^{(k)} \s(\vv{1}_{i,(k)}, \vv{2}_{j,(k)})
\end{equation}

\section{Results and Discussion}

\begin{table}[t]
	\centering
	\caption{Weakly Supervised Model Results on STS-Benchmark Dataset.}\label{tb:weak-supervise}
		\vskip-1em
	{\scriptsize
	\begin{tabular}{|l|ll|}
		\hline
		Weakly Supervised Model                                                          & Dev  & Test \\
		\hline
		InferSent (bi-LSTM trained on SNLI)~\cite{DBLP:conf/emnlp/ConneauKSBB17}         & 80.1 & 75.8 \\
		Sent2vec~\cite{DBLP:conf/naacl/PagliardiniGJ18}                                  & 78.7 & 75.5 \\
		Conversation response prediction + SNLI~\cite{DBLP:conf/rep4nlp/YangYCKCPGSSK18} & 81.4 & 78.2 \\
		SIF on Glove vectors~\cite{DBLP:conf/iclr/AroraLM17}                             & 80.1 & 72.0   \\
		GRAN (uses SimpWiki)~\cite{DBLP:conf/acl/WietingG17}                             & 81.8 & 76.4 \\
		Unsupervised SIF + ParaNMT vectors~\cite{DBLP:conf/rep4nlp/EthayarajhH18}       & 84.2 & 79.5 \\
		GEM~\cite{yang2019parameter} & 83.5 & 78.4 \\
		\hline
		ROTS+ binary tree (ours) & 84.4 & 80.0 \\
		ROTS+ dependency tree (ours)                                                  & \textbf{84.6} & \textbf{80.6} \\ \hline
	\end{tabular}
	}
\end{table}

\begin{table}
	\centering
	\caption{Detailed Comparisons with Similar Unsupervised Approaches on 20 STS Datasets}\label{tb:fg}
	\vskip-1em
	{\tiny
	\begin{tabular}{|l|l|cccc|c|}
		\hline
		Model Type& Senrence Similarity &     STS12     &     STS13     &     STS14     &     STS15     &      AVE      \\ \hline
		\multirow{3}{*}{\shortstack[l]{OT based}}& WMD\shortcite{DBLP:conf/icml/KusnerSKW15}        &     60.6      &     54.5      &     65.5      &     61.8      &     60.6      \\
		&WME\shortcite{DBLP:conf/emnlp/WuYXXBCRW18}        &     62.8      &     65.3      &      68       &     64.2      &    65.1     \\
		        &CoMB\shortcite{DBLP:conf/iclr/SinghHDJ19}        &     57.9      &     64.2      &     70.3      &     73.1      &    66.4     \\
        \hline
		\multirow{3}{*}{\shortstack[l]{Weighted\\ average}} &SIF\shortcite{DBLP:conf/iclr/AroraLM17}         &     59.5      &     61.8      &     73.5      &     76.3      &    67.8     \\
		&uSIF\shortcite{DBLP:conf/rep4nlp/EthayarajhH18}     &     65.8      & 66.1 &     78.4      &     79.0      &     72.3      \\
		&DynaMax\shortcite{DBLP:conf/iclr/ZhelezniakSSMFH19} &  66.0 &     65.7      &     75.9     &     \textbf{80.1}      &     72.0      \\ \hline
		\multirow{2}{*}{Ours} &ROTS + binary tree                        &    \textbf{68.3}      &      66.0       & \textbf{78.7} &  79.5    & 73.1 \\
		& ROTS + dependency tree&67.5&\textbf{66.4}&\textbf{78.7}& 80.0 & \textbf{73.2}\\\hline
	\end{tabular}}
		\vskip-1em
\end{table}

\subsection{Semantic Textual Similarity Tasks and Baselines}
In the results part, we compare our ROTS model with others on the semantic textual similarity tasks. 
Firstly, Our major results are based on STS-benchmark dataset~\cite{DBLP:journals/corr/abs-1708-00055}. STS-benchmark fuses many STS datasets from year 2012 - 2017 and provides a standard setup for training, dev and test on three selected genres (news, captions, forums). We compare the results on dev and test set with the leaderboard on benchmark dataset~\footnote{http://ixa2.si.ehu.es/stswiki/index.php/STSbenchmark}.
Secondly, for more detailed comparisons, we also include SemEval semantic texual similarity tasks (2012-2015)~\cite{DBLP:conf/semeval/AgirreCDG12,DBLP:conf/starsem/AgirreCDGG13,DBLP:conf/semeval/AgirreBCCDGGMRW14,DBLP:conf/semeval/AgirreBCCDGGLMM15}. There are 20 datasets in total. 
Since our model is related to OT distance and cosine similarity of WAWV, we compare with OT based models~\cite{DBLP:conf/icml/KusnerSKW15} , kernels\cite{DBLP:conf/emnlp/WuYXXBCRW18} and barycenters~\cite{DBLP:conf/iclr/SinghHDJ19},  WAWV based models~\cite{DBLP:conf/rep4nlp/EthayarajhH18,DBLP:conf/iclr/AroraLM17} and another recent model improves the WAWV~\cite{DBLP:conf/iclr/ZhelezniakSSMFH19}.
Finally, we conduct the ablation study again on STS-benchmark dataset to demonstrate how ROTS works.
The scores for all semantic textual similarity datasets are the Pearson's correlation coefficient times 100.

\begin{table*}[t]
	\centering
	\caption{Ablation Study for 4 Aspects of ROTS. Bold values indicates the best in the row}\label{tb:ablation1}
	\vskip-1em
	\begin{tabular}{|c|c|ccccc|c|c|}
		\hline
        Comments & Dataset & d=1& d=2 & d=3 & d=4 & d=5 & ROTS& uSIF\\
        \hline
		\multirow{2}{*}{ROTS + ParaNMT Vectors + Dependency Tree}&STSB dev & 84.5 & 84.6 & 84.6 & 84.5 & 84.4 & \textbf{84.6} & 84.2\\
		&STSB test & 80.5 & 80.6  & 80.6 & 80.5 & 80.4 & \textbf{80.6} & 79.5\\
		\hline
		\multirow{2}{*}{ROTS + ParaMNT Vectors + Binary Tree}&STSB dev & 84.2 & 84.3 & 84.3 & 84.3 & 84.3 & \textbf{84.4} & 84.2\\
		&STSB test & 79.7 & 79.9 & 80.0 & 80.0 & 80.0 & \textbf{80.0} & 79.5\\
		\hline
		\multirow{2}{*}{ROTS + GloVe Vectors + Dependency Tree}&STSB dev & 78.5 & 78.9 & 79.2 & 79.5 & 79.7 & \textbf{79.3} & 78.4\\
		&STSB test & 72.0 & 72.5 & 73.0 & 73.3 & 73.6 & \textbf{73.0} & 70.3\\
		\hline
		\multirow{2}{*}{ROTS + PSL Vectors + Dependency Tree}&STSB dev & 80.9 & 80.9 & 80.8 & 80.7 & 80.6 & 80.9 & \textbf{81.0}\\
		&STSB test & 75.4 & 75.5 & 75.6 & 75.6 & 75.5 & \textbf{75.6} & 74.4\\
		\hline
		\multirow{2}{*}{Independent OT + ParaMNT Vectors + Dependency Tree}&STSB dev & 76.6 & 74.8 & 77.1 & 72.1 & 62.5 & 75.6 & \textbf{84.2}\\
		&STSB test & 69.4  & 68.1 & 68.2 & 59.0 & 50.4 & 65.8 & \textbf{79.5}\\
		\hline
		\multirow{2}{*}{\shortstack[l]{ROT without modifications on coefficients $C$ and $a, b$ weights \\ + ParaMNT Vectors + Dependency Tree}}&STSB dev & 69.8 & 69.0 & 69.0 & 67.9 & 66.1 & 74.8 & \textbf{84.2}\\
		&STSB test & 64.8  & 64.0 & 64.7 & 63.8 & 61.9 & 71.0 & \textbf{79.5}\\
		\hline
	\end{tabular}
	\vskip-1em
\end{table*}

\subsection{Implimentation Details}

The ROTS requires validation of the hyperparameters. For simplicity, we restrict the depth of ROTS model to be 5, and the $ \beta_k $ for different depths are exactly the same $ \beta $. The ROTS hyperparameter $ \beta $ is selected from the dev set of STS Benchmark. Finally, we choose $ \beta=10 $ for all results.
As has stated, our ROTS framework admits different hierarchical structures. Here we use two kinds of tree structures. The one is binary tree that constructed from a top-down approach (described in Section~\ref{sec:tree} like Figure~\ref{fig:hots}). The other is the dependency tree generated from SpaCy~\cite{spacy2} with \texttt{en\_core\_web\_sm} model.
In addition, for the first level of ROTS model, we directly solve optimal transport.
The weight scheme is chosen to be uSIF weights~\cite{DBLP:conf/rep4nlp/EthayarajhH18}) which requires only statistical information.
The pre-trained word vectors includes GloVe~\cite{DBLP:conf/emnlp/PenningtonSM14}, PSL~\cite{DBLP:journals/tacl/WietingBGL15} and ParaNMT word vectors~\cite{DBLP:conf/acl/GimpelW18}). 
We preprocess our word embeddings by removing the main components like uSIF~\cite{DBLP:conf/rep4nlp/EthayarajhH18}.

\subsection{Major Results on the STS Benchmark}
The benchmark comparisons are focused on weakly supervised models with no parameters to train but only hyper-parameters to select. 
A detailed results are listed in Table~\ref{tb:weak-supervise}. The results that we compared are gathered from either the leaderboard in STS-Benchmark website or from directly the best reported models in the paper~\cite{yang2019parameter}.
The presented ROTS results are based on uSIF weights~\cite{DBLP:conf/rep4nlp/EthayarajhH18} and ParaNMT word vectors~\cite{DBLP:conf/acl/GimpelW18} with both binary tree and dependency tree.

We could see in Table~\ref{tb:weak-supervise} that both ROTS has better performance than all previous weakly supervised models. 
The ROTS model gets much more improvement compared to other models proposed in year 2019.
ROTS with dependency parsing tree increased about one point than previous state-of-the-art uSIF results. 
This means that hierarchical prior takes significant effects.

\subsection{Detailed Study Results over 20 STS datasets}
We could see from Table~\ref{tb:fg} that the OT based approaches are consistently worse than weighted average based approaches. 
Due to the difference of implementation, choice of weights and word embeddings, scores mentioned in Table~\ref{tb:fg} are highest scores selected from their original papers.
The previous latest DynaMax~\cite{DBLP:conf/iclr/ZhelezniakSSMFH19} variation has compatible performance than uSIF in individual scores and the average score. 
While our HOT approaches almost steadily outperformed uSIF and has best average scores. 
Notebly, best DynaMax model also employs \textbf{the same uSIF weights as well as ParaMNT word vectors}. So the improvement of ROTS over uSIF and DynaMax is fair and clear.

\subsection{Ablation Study}
We investigate three aspects of ROTS: 1) What are the similarities from each level; 2) What are the effects of different word vectors and tree structures; 3) Is it necessary for ROTS to pass prior information from levels using ROT? and 4) The effect of the adjustments by semantic insights mentioned in Section~\ref{sec:insights}.
In detailed study over 20 datasets, we have demonstrate the superiority of uSIF~\cite{DBLP:conf/rep4nlp/EthayarajhH18} over all other models. So we only reports the results of uSIF to make a comparison when we changes the conditions.

In Table~\ref{tb:ablation1}, we could see from the first two cases that the performance generally gets better when going deeper. The dependency tree's score is higher than that from binary tree, while it slightly decrease after 3-th level.
In the first four cases that we could see the results by changing the word vectors and tree structures. We conclude that almost at every case our approach consistently outperformed uSIF and ParaMNT and Dependency Tree is the best combination.
In STS-Benchmark test set, ROTS outperformed uSIF about 1.4 in average of those four different cases.
The last two cases demonstrated the necessity of both recursive prior information passing and semantic insights by Section~\ref{sec:insights}.
Without any of those, the results is much worse than uSIF.

\section{Conclusion}

In this paper, We present ROT framework to incorporate the structural information into the comparison of distributional data. We develop ROTS sentence similarity with valuable semantic insights from the connection that we established between cosine similarity of WAWV and OT. Extensive experiments shows that our approach consistently improves previous state-of-the-art models in varying conditions. Calculation of ROT only requires $O(n)$ time complexity. Compared to classic optimal transport that needs $O(n^2)$, ROT could be scaled up to much larger cases.
Finally, we emphasize that the ROT framework applies to more general distributional data comparison where structural information is important.

\bibliographystyle{named}
\bibliography{ref}

\end{document}